\documentclass{article} %
\usepackage{iclr2024_conference,times}

\usepackage{amsmath,amsfonts,bm}

\def\eqref#1{equation~\ref{#1}}

\def\1{\bm{1}}

\DeclareMathAlphabet{\mathsfit}{\encodingdefault}{\sfdefault}{m}{sl}
\SetMathAlphabet{\mathsfit}{bold}{\encodingdefault}{\sfdefault}{bx}{n}

\usepackage{hyperref}
\usepackage{url}

\usepackage{times}
\usepackage{latexsym}

\usepackage[T1]{fontenc}

\usepackage[utf8]{inputenc}

\usepackage{inconsolata}
\usepackage[frozencache,cachedir=_minted-output]{minted}
\setminted{
fontsize=\scriptsize,
}
\usepackage{adjustbox}

\usepackage[utf8]{inputenc} %
\usepackage[T1]{fontenc}    %
\usepackage{hyperref}       %
\usepackage{url}            %
\usepackage{booktabs}       %
\usepackage{amsfonts}       %
\usepackage{nicefrac}       %
\usepackage{microtype}      %
\usepackage{xcolor}         %
\usepackage{graphicx}
\usepackage{listings}
\usepackage{amsmath}
\usepackage{amssymb}
\usepackage{cleveref}
\usepackage{multicol}
\usepackage{multirow}
\usepackage{mathtools}
\usepackage{makecell}
\usepackage{cprotect}
\usepackage{varwidth}
\usepackage{wrapfig}
\usepackage{pifont}
\usepackage{caption}
\usepackage{enumitem}
\usepackage{tikz}
\usepackage{pgfplots}\pgfplotsset{compat=1.9}
\usepackage{colortbl}

\usepackage{algorithm}
\usepackage[noend]{algpseudocode}

\newcommand{\lajan}[1]{}

\newcommand{\starcoder}{StarCoder~}
\newcommand{\codegen}{CodeGen~}
\newcommand{\schema}[1]{\emph{#1}}
\newcommand{\act}{\mathcal{A}}
\newcommand{\obs}{\mathcal{O}}
\newcommand{\ctxt}{\tau_{<t}}

\newcommand{\dataset}{\mathcal{D}}
\newcommand{\precond}{\mathcal{F}}
\newcommand{\precfn}[2]{g^\text{#1}_{#2}}
\newcommand{\testset}{\mathcal{D}^\text{test}}

\newcommand{\exps}{\mathop{\mathbb{E}}}

\newcommand{\actfn}{f_a}
\newcommand{\alfworld}{ALFworld}

\newcommand{\cmark}{\ding{51}}%
\newcommand{\xmark}{\ding{55}}%

\newcommand{\cutfigureabove}{\vspace{0em}}
\newcommand{\cutfigurebelow}{\vspace{-0.4em}}
\newcommand{\cuttableabove}{\vspace{0em}}
\newcommand{\cuttablebelow}{\vspace{-0.4em}}

\definecolor{codegreen}{rgb}{0,0.6,0}
\definecolor{codegray}{rgb}{0.5,0.5,0.5}
\definecolor{codepurple}{rgb}{0.58,0,0.82}
\definecolor{backcolour}{rgb}{0.95,0.95,0.92}

\lstdefinestyle{mystyle}{
    backgroundcolor=\color{backcolour},   
    commentstyle=\color{codegreen},
    keywordstyle=\color{magenta},
    stringstyle=\color{codepurple},
    basicstyle=\ttfamily\scriptsize,
    breakatwhitespace=false,         
    breaklines=true,                 
    captionpos=b,                    
    keepspaces=true,                 
    showspaces=false,                
    showstringspaces=false,
    showtabs=false,                  
    tabsize=2
}

\title{Code Models are Zero-shot Precondition \\ Reasoners}

\author{
\hspace{-0.25em}Lajanugen Logeswaran$^1$,
Sungryull Sohn$^{1}$,
Yiwei Lyu$^{2}$,
Anthony Zhe Liu$^2$,
Dong-Ki Kim$^1$, \\
\textbf{Dongsub Shim}$^1$,
\textbf{Moontae Lee}$^1$,
\textbf{Honglak Lee}$^{1,2}$ \\[0.2em]
$^1$LG AI Research \hspace{1em} $^2$University of Michigan, Ann Arbor
}

\iclrfinalcopy %
\begin{document}

\maketitle
\begin{abstract}

One of the fundamental skills required for an agent acting in an environment to complete tasks is the ability to understand what actions are plausible at any given point.
This work explores a novel use of code representations to reason about action preconditions for sequential decision making tasks.
Code representations offer the flexibility to model procedural activities and associated constraints as well as the ability to execute and verify constraint satisfaction.
Leveraging code representations, we extract action preconditions from demonstration trajectories in a zero-shot manner using pre-trained code models.
Given these extracted preconditions, we propose a precondition-aware action sampling strategy that ensures actions predicted by a policy are consistent with preconditions.
We demonstrate that the proposed approach enhances the performance of few-shot policy learning approaches across task-oriented dialog and embodied textworld benchmarks.

\end{abstract}

\section{Introduction}

A key capability for learning an optimal agent policy in sequential decision making settings is understanding the plausibility of different actions.
For instance, a dialog agent recommending restaurants needs to have basic information like location and type of food in order to look up its database for potential options.
Understanding the necessary conditions for performing an action (e.g., location and type of food are necessary for the database lookup action) is referred to as \emph{precondition inference} or \emph{affordance learning} in the literature \citep{ahn2022can,sohn2020meta}.
Knowledge about preconditions has broad implications for policy learning and safety applications.

Few-shot prompted large language models (LLMs) have demonstrated strong capabilities for task planning \citep{logeswaran2022few,huang2022language,micheli2021language,ahn2022can}.
However, they lack a systematic mechanism to reason about action preconditions. %
Some approaches attempt to teach LLMs preconditions by providing examples in the form of code assertions \citep{singh2022progprompt,liang2022code} or natural language rationales \citep{huang2022inner,yao2022react}.
Preconditions or rationales for predicted actions are generated dynamically on-the-fly the during inference.
However, it is non-trivial to verify whether these preconditions or rationales are adequate as they are dynamically generated.
It is further difficult to guarantee with certainty that predicted actions will be consistent with the preconditions or rationales.

Taking inspiration from classical AI planning literature \citep{aeronautiques1998pddl,fikes1993strips} and recent work that use code representations for reasoning problems, we leverage code representation to reason about preconditions. %
Programs are a natural formalism to model event sequences and offer the flexibility to express dependency constraints between events, such as in the form of assertions \citep{liang2022code,singh2022progprompt}.
Verifying that a program meets the specifications dictated by the assertions amounts to simply executing the program and verifying that the program ran successfully. 
The ability to execute and verify constraint satisfaction is a key benefit of code as a representations compared to alternative representations such as natural language.
Representing preconditions in the form of procedural statements in code further provides transparency, controllability and better generalization to unseen scenarios compared to alternative representations of affordance such as neural network functions. %
Finally, this also enables us to exploit strong priors captured by code understanding models for policy learning problems.

\begin{figure*}[!t]
\centering
\includegraphics[width=1.0\textwidth,trim={0 4.5em 12em 0em},clip]{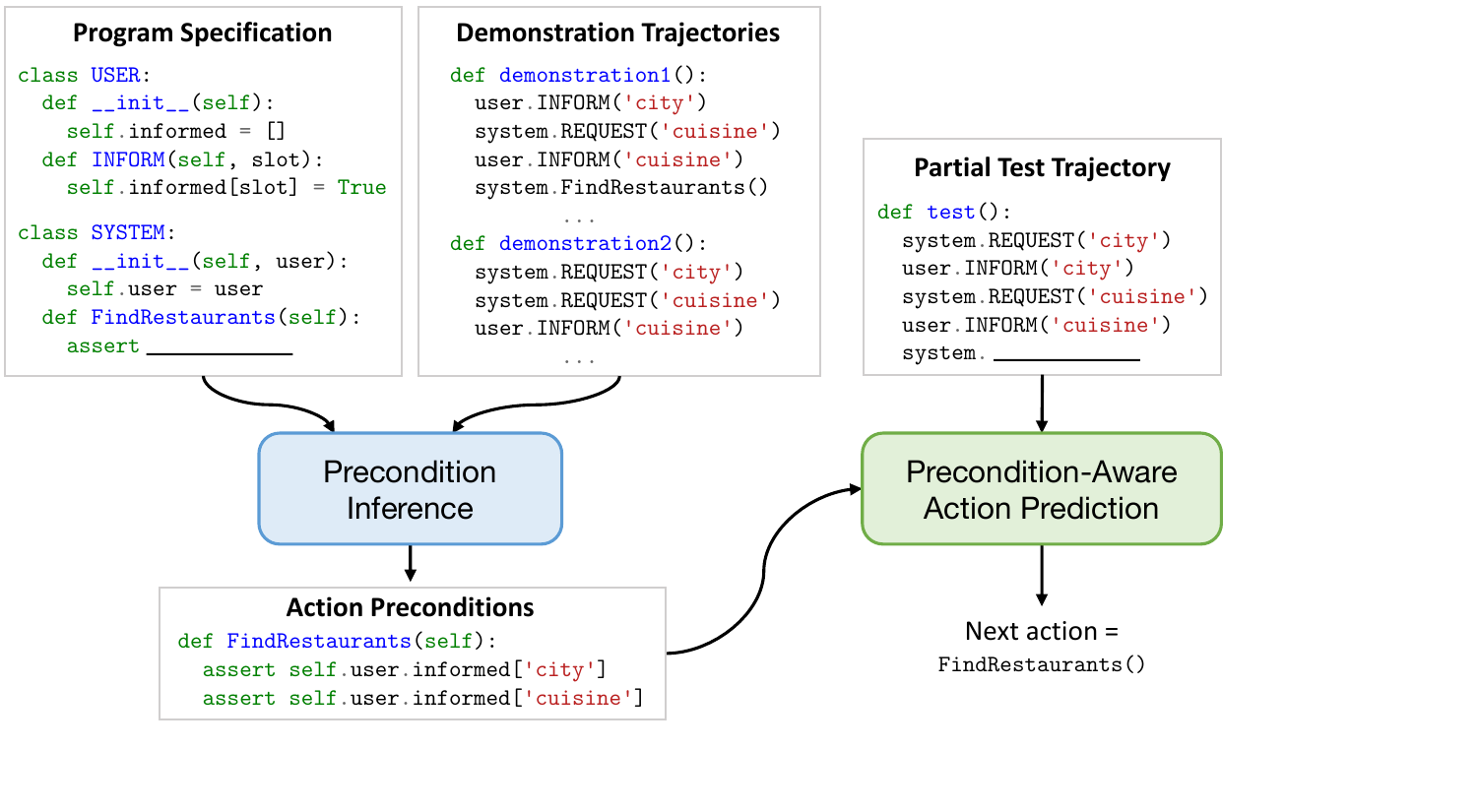}
\cutfigureabove
\caption{
\textbf{Approach Overview}:
We propose a framework to extract action preconditions from demonstration trajectories by leveraging code representation and pre-trained code models. 
Extracted preconditions are used to construct an agent policy that predicts actions consistent with preconditions. 
}
\label{fig:example}
\cutfigurebelow
\end{figure*}

Armed with a code representation, we formulate precondition inference as a code completion problem and propose a zero-shot approach to infer preconditions of actions given demonstration trajectories leveraging pre-trained code models.
Given these inferred preconditions, we propose a precondition-aware action prediction strategy that ensures actions predicted by a policy are consistent with the preconditions.
We present extensive analysis and ablations that show the impact of different components of our approach.

In summary, we make the following contributions in this work.
\begin{itemize}[leftmargin=1em,topsep=2pt,noitemsep]
\item We propose a new approach that leverages program representations and pre-trained code models to extract action preconditions from expert demonstrations alone in a zero-shot manner.
\item Combining inferred preconditions with a precondition-aware action prediction strategy, we demonstrate that the proposed framework leads to better agent policies compared to baselines on task-oriented dialog and embodied textworld benchmarks. 

\end{itemize}

\section{Problem Setting}
\label{sec:problem}

We consider a sequential decision making setting where the agent receives a sequence of observations $o_i\in\obs$ and performs an action $a_i\in\act$, where we assume discrete observation and action spaces $\obs,\act$.
The agent's trajectory can be represented as a sequence of observations and actions $\tau = (o_1, a_1, o_2, a_2, \ldots, o_n, a_n)$.
We consider a few-shot learning setting where we are given demonstrations $\dataset = \{\tau^1, \ldots, \tau^n\}$. 

Let $\ctxt = (o_{1:t}, a_{1:t-1})$ represent the history of observations and actions.
The goal of learning is two-fold.
First, for each action $a$ we intend to estimate a precondition function $\precfn{}{a}(\ctxt) \in \{0, 1\}$ that informs whether the action is plausible in a given context $\ctxt$ (i.e., $\precfn{}{a}(\ctxt) = 0$ represents the action is implausible and $\precfn{}{a}(\ctxt) = 1$ represents the action is plausible).
Given these inferred preconditions $\mathcal{G} = \{g_a | a \in \mathcal{A} \}$, our second goal is to estimate a policy $\pi(a_t|\ctxt, \dataset, \mathcal{G})$ that predicts the next action given $\ctxt$.
The overall goal is to construct a policy $\pi$ that generalizes to a set of test trajectories $\testset$: maximize $\exps_{\tau \in \testset} [\text{log } \pi(a_t | \ctxt, \dataset, \mathcal{G})]$.

\section{Approach}
\label{sec:approach}

We first discuss the code representation in \Cref{sec:programs}.
We then describe our approach to precondition inference and action prediction problems in \Cref{sec:precond,sec:policy}, respectively.
See \Cref{fig:example} for an overview of our approach.

\subsection{Representing Agent Trajectories as Programs}
\label{sec:programs}

We represent the agent's interaction with the environment as a program.
We represent every action and observation as one or more function calls that modify the state of a predefined set of variables $v$ that capture a summary of the agent's experience (e.g., observations $o_{1:t}$ and actions $a_{1:t-1}$).\footnote{For instance, these could be boolean variables analogous to predicates that represent state in PDDL planning.}
Such a representation exists since we assume discrete observation and action spaces (for instance, we can define a separate function for each string in $\obs,\act$).
Assume we have defined a set of functions $\precond_\obs, \precond_\act$ corresponding to observations $\obs$ and actions $\act$ respectively.
Given this representation, a trajectory $\tau$ can be viewed as a program which consists of a sequence of function calls.\footnote{We interchangeably refer to actions/observations as functions and trajectories as programs in the paper.}
We present an example program in \Cref{fig:example} and further examples in the Appendix.

\subsection{Precondition Inference}
\label{sec:precond}

Recall that the precondition function $\precfn{}{a}(\ctxt) \in \{0, 1\}$ informs whether an action $a$ is plausible in a given context $\ctxt$. %
Equivalently, for the corresponding function in the program representation $\actfn \in \precond_\act$, we seek to identify assertion statements in terms of variables $v$ (which represent a summary of $\ctxt$) and the function arguments of $f_a$. 
(For example in \Cref{fig:example}, the precondition for the action {\small \texttt{FindRestaurants}} is identified as {\small \texttt{assert user.informed[`city'] and user.informed[`cuisine']}}).

We predict the preconditions for each action independently of other actions and the process described below is repeated for each action $a \in \act$.\footnote{Jointly reasoning about preconditions for all functions in $\precond_\act$ can be interesting and is left to future work.}
Our approach to precondition inference consists of the following steps: candidate generation, validation and ranking.
We detail these steps next (See \Cref{appsec:precondition inference} for an illustration and \Cref{appsec:precond inf ex} for example outputs from each component of the pipeline).

\paragraph{Candidate Generation.}
Given demonstration trajectories $\dataset$ we first generate candidate preconditions by prompting a pre-trained code generation model. 
The prompt consists of (i) a demonstration $\tau\in\dataset$ (ii) definitions of functions $\precond_\obs$ and (iii) definition of function $\actfn$ with the \texttt{assert} keyword in its body.
The \texttt{assert} prefix forces the model to generate assertion statements (as opposed to arbitrary code).
We vary the demonstration program and sample multiple precondition candidates for each demonstration to come up with a pool of plausible candidates $\mathcal{H}^\text{initial}_a$.
The pre-trained model is expected to understand what are appropriate contexts in which a function can be used and use this understanding to come up with assertion statements.

Although pretrained models have strong priors about appropriate assert statements, the above process has some limitations.
First, generated statements may not be meaningful due to syntax errors or other deficiencies in the generated expressions.
Second, the candidates are obtained based on static analysis of the program alone and the model has to implicitly reason about execution and program state in order to predict accurate preconditions.

\paragraph{Candidate Validation.}
One of the key advantages of a program representation as opposed to alternative representations such as natural language is the ability to execute.
We augment the above candidate generation approach with a verification approach where each candidate is vetted for validity and consistency with the data.
Given a candidate assertion, we verify whether each of the demonstration programs can run successfully (e.g,. replacing the function body with a candidate assertion and executing a demonstration program). %
Candidates which led to execution failures are discarded.
The remaining candidates $\mathcal{H}^\text{valid}_a$ are thus valid and consistent with the data. 

Many of the assertions generated will not be useful in practice.
For example, trivial assertions such as `\texttt{assert True}' do not convey any useful information about instances where the function is applicable.
Although all the candidate assertions at this point are valid, they may be \emph{sub-optimal} for the purpose of constructing a policy.
We seek assertions that help discriminate situations where the function is applicable.
We propose a ranking mechanism to identify the most discriminative assertions.

\paragraph{Candidate Ranking.}

We seek to identify a small set of precondition/assertion candidates from $\mathcal{H}^\text{valid}$ useful for constructing an agent policy.
We begin with the observation that if candidates $h_1, h_2$ are such that $h_2$ is satisfied whenever $h_1$ is satisfied (i.e., $h_1 \Rightarrow h_2$), $h_1$ is more desirable as it is more discriminative of the contexts where action $a$ is applicable (and hence leads to a better policy).
As we discuss in the experiments, this assumption leads to high precision solutions and compromises recall.
We leave alternate ranking criteria to future work.

However, verifying the above property ($h_1 \Rightarrow h_2$) for given precondition candidates is generally intractable.
We thus consider an approximation where we examine if the property is satisfied in scenarios that appear in the demonstration trajectories.
Consider the precondition function $g_a({}\cdot{}; h)$ which assumes $h$ to be the precondition of action $a$.
Define $C_a^h = \{(i, j) | g_a(\tau^i_{<j}; h) = 1, \tau^i \in \dataset\}$ to be the set of instances $(i, j)$ where precondition $h$ is satisfied at time-step $j$ of demonstration trajectory $\tau^i$.
We use $C_a^{h_1} \subseteq C_a^{h_2}$ as a proxy to determine whether $h_1 \Rightarrow h_2$.

We define the optimal set of assertion statements as in \Cref{eq:hopt}.
When multiple equivalent candidates exist, we choose one random representative to retain in set $\mathcal{H}^\text{opt}_a$.
Note that the conjunction of assertions in $\mathcal{H}^\text{opt}_a$ is equivalent to that of $\mathcal{H}^\text{valid}_a$.
However, identifying a small set of assertion statements is beneficial both for interpretability and for providing as a prompt for models which have limited context lengths.
\begin{equation}
\mathcal{H}^\text{opt}_a = \{h\in \mathcal{H}^\text{valid}_a \mid \nexists h' \in \mathcal{H}^\text{valid}_a \text{ s.t. } C_a^{h'} \subset C_a^h \}
\label{eq:hopt}
\end{equation}

\subsection{Precondition-aware Action Prediction}
\label{sec:policy}

We pose action prediction as a code completion problem where given a partial agent trajectory, a code model is tasked with suggesting possible next actions (as functions from $\precond_\act$, with appropriate arguments).
Given any policy which predicts the next action given past actions and observations, we consider a simple approach to augment the policy with precondition knowledge. 
Next action candidates are sampled from the policy until an action consistent with the preconditions is found or a maximum number of attempts is exceeded.
The first attempt uses greedy sampling and subsequent attempts resort to random sampling.
If random sampling does not yield an action consistent with preconditions, we default to the greedy action.
We sample an action by generating tokens until a newline token is encountered.
See \Cref{appsec:action prediction} for the algorithm pseudocode.

\section{Experiments}

We attempt to answer the following key questions in our evaluation: 
(i) Is it possible to extract information about action preconditions from demonstrations of agent behavior?
(ii) Is such inferred precondition information useful for building better agent policies?

We use the python programming language as the code representation in our experiments due to its simplicity and popularity, as well as the recent release of code models that primarily focus on python.
We use the \codegen 2B \citep{nijkamp2022codegen} and \starcoder 16B \citep{li2023starcoder} open-source pre-trained code models in our experiments.
Specifically, we use the versions of these models which were further pre-trained on python code after initial pre-training on many programming languages.

\subsection{Benchmarks}
\paragraph{Task Oriented Dialog Benchmark.}
The decision making component of a task-oriented dialog system is a dialog manager which takes as input a sequence of utterances represented as dialog acts and predicts the next action.
In this setting, a user and a system (agent) take turns to speak and user utterances constitute the agent's observations.
We use the SGD dataset \citep{rastogi2020towards} in our experiments.
There are 11 user acts and 11 system acts defined in the dataset, each of which takes either no argument, a single slot argument or a single intent argument.
We define analogous functions $\precond_\obs$ and $\precond_\act$ corresponding to each of these acts.
We take an object-oriented approach and group these functions respectively under a user class and a system class.
We define variables $v$ corresponding to each user action that records whether the action was performed (e.g., \texttt{informed\_slot[]}, \texttt{requested\_slot[]} in \Cref{fig:example}).
We experiment with 10 domains (schemas) from the SGD dataset. %
10 instances are used as demonstrations and 50 instances for testing.
For evaluation purposes we manually define the ground-truth preconditions for each system action based on prior knowledge about the dialog acts.
Note that these are only used for evaluation and no supervision is provided to the model about ground-truth preconditions.

\paragraph{Embodied Textworld.}
We experiment with the \alfworld~embodied textworld benchmark \citep{shridhar2020alfworld} which involves an agent interacting with an environment to perform object interaction tasks.
The observations and actions are natural language statements and the agent is expected to perform a task specified using a natural language instruction (e.g., \emph{move the keys to the table}).
There are 9 types of actions which include interaction and navigation actions.
Each of these action types take an object argument and/or a receptacle argument and we define analogous functions $\precond_\act$. 
In addition, we define auxiliary functions $\precond_\obs$ for adding/removing an object from the agent's inventory and updating the set of objects visible to the agent.
We define three variables $v$ that summarize the agent's experience: the set of visible objects, the agent's inventory and states of objects in the environment (e.g., open/closed).
The dataset has 6 task types.
We use 2 instances from each task type as demonstrations and the standard test set of the benchmark for testing.
The benchmark provides ground-truth preconditions for the actions in a PDDL (Planning Domain Definition Language) representation, which we convert to python assertions for evaluation.
We provide the complete program specification for these benchmarks in \Cref{appsec:spec}.

\begin{table*}[!t]
\small
\centering
\begin{tabular}{l l c c}
\toprule
Ground-truth precondition & Predicted precondition & Prec & Rec \\ %
\hline
\\[-0.8em]
\begin{minipage}[t]{.4\textwidth}
\begin{minted}{python}
def INFORM(self, slot):                            
  assert self.user.requested_slot[slot]
\end{minted}
\end{minipage}
& 
\begin{minipage}[t]{.4\textwidth}
\begin{minted}{python}
def INFORM(self, slot):                            
  assert slot in self.user.requested_slot               
\end{minted}
\end{minipage}
& 1 & 1 \\[1.0em] %
\begin{minipage}[t]{.4\textwidth}
\begin{minted}{python}
def REQUEST(self, slot):                           
  assert not self.user.informed_slot[slot]
\end{minted}
\end{minipage}
& 
\begin{minipage}[t]{.4\textwidth}
\begin{minted}{python}
def REQUEST(self, slot):                           
  assert self.user.informed_slot[slot] == False    
\end{minted}
\end{minipage}
& 1 & 1 \\[1.0em] %
\begin{minipage}[t]{.4\textwidth}
\begin{minted}{python}
def GOODBYE(self):                                 
  assert self.user.no_more            
\end{minted}
\end{minipage}
& 
\begin{minipage}[t]{.4\textwidth}
\begin{minted}{python}
def GOODBYE(self):                                 
  assert self.user.no_more            
\end{minted}
\end{minipage}
& 1 & 1 \\[1.0em] %
\begin{minipage}[t]{.4\textwidth}
\begin{minted}{python}
def OFFER(self, slot):                             
  assert self.query_success
\end{minted}
\end{minipage}
& 
\begin{minipage}[t]{.4\textwidth}
\begin{minted}{python}
def OFFER(self, slot):                             
  assert self.query_success == True            
\end{minted}
\end{minipage}
& 1 & 1 \\[1.0em] %
\begin{minipage}[t]{.4\textwidth}
\begin{minted}{python}
def INFORM_COUNT(self):                            
  assert self.query_success
\end{minted}
\end{minipage}
& 
\begin{minipage}[t]{.4\textwidth}
\begin{minted}{python}
def INFORM_COUNT(self):                            
  assert self.query_success == True            
\end{minted}
\end{minipage}
& 1 & 1 \\[1.0em] %
\begin{minipage}[t]{.4\textwidth}
\begin{minted}{python}
def OFFER_INTENT(self, intent):                    
  assert self.user.selected
\end{minted}
\end{minipage}
& 
\begin{minipage}[t]{.4\textwidth}
\begin{minted}{python}
def OFFER_INTENT(self, intent):                    
  assert self.query_success == True            
\end{minted}
\end{minipage}
& 0.49 & 1 \\[1.0em] %
\begin{minipage}[t]{.4\textwidth}
\begin{minted}{python}
def CONFIRM(self, slot):                           
  assert self.user.informed_slot['to_location']   
  assert self.user.informed_slot['from_location']  
  assert self.user.informed_slot['leaving_date']   
  assert self.user.informed_slot['travelers']  
\end{minted}
\end{minipage}
& 
\begin{minipage}[t]{.4\textwidth}
\begin{minted}{python}
def CONFIRM(self, slot):                           
  assert self.user.selected
  assert slot != 'travelers' \
    or self.user.informed_slot[slot]

\end{minted}
\end{minipage}
& 0.96 & 0.7 \\[3.7em] %
\begin{minipage}[t]{.4\textwidth}
\begin{minted}{python}
def NOTIFY_SUCCESS(self):                          
  assert self.user.query_success
 
\end{minted}
\end{minipage}
& 
\begin{minipage}[t]{.4\textwidth}
\begin{minted}{python}
def NOTIFY_SUCCESS(self):                          
  assert self.user.selected            
  assert not self.user.no_more            
\end{minted}
\end{minipage}
& 1 & 0.44 \\[2.0em] %
\begin{minipage}[t]{.4\textwidth}
\begin{minted}{python}
def REQ_MORE(self):                                
  assert self.user.selected or self.user.no_more
\end{minted}
\end{minipage}
& 
\begin{minipage}[t]{.4\textwidth}
\begin{minted}{python}
def REQ_MORE(self):                                
  assert self.user.selected            
\end{minted}
\end{minipage}
& 1 & 1 \\[1.0em] %
\begin{minipage}[t]{.4\textwidth}
\begin{minted}{python}
def FindBus(self):                                 
  assert self.user.informed_slot['to_location']   
  assert self.user.informed_slot['from_location']  
  assert self.user.informed_slot['leaving_date']  

\end{minted}
\end{minipage}
& 
\begin{minipage}[t]{.4\textwidth}
\begin{minted}{python}
def FindBus(self):                                 
  assert not self.user.no_more            
  assert self.user.informed_intent['FindBus']      
  assert self.user.informed_slot['leaving_date']   
  assert self.user.informed_slot['from_location']  
\end{minted}
\end{minipage}
& 0.99 & 0.89 \\[3.7em] %
\begin{minipage}[t]{.4\textwidth}
\begin{minted}{python}
def BuyBusTicket(self):                            
  assert self.user.affirmed
\end{minted}
\end{minipage}
& 
\begin{minipage}[t]{.4\textwidth}
\begin{minted}{python}
def BuyBusTicket(self):                            
  assert self.user.affirmed
\end{minted}
\end{minipage}
& 1 & 1 \\[0.8em] %
\arrayrulecolor{black}
\bottomrule
\end{tabular}
\caption{Ground-truth and predicted preconditions for actions in the \schema{Buses} domain of the SGD benchmark, along with Precision, Recall scores for predictions.}
\cuttableabove
\label{table:qualitative}
\cuttablebelow
\end{table*}

\definecolor{light-gray}{gray}{0.96}

\subsection{Precondition Inference}

\paragraph{Metrics.}

\newcommand{\cpred}{C_a^\text{pred}}
\newcommand{\cgt}{C_a^\text{gt}}

Recall that a precondition $g_a(\ctxt) \in \{0, 1\}$ informs whether an action $a \in \mathcal{A}$ is plausible given context $\ctxt = (o_{1:t}, a_{1:t-1})$. %
Define $C_a = \{(i, j) | g_a(\tau^i_{<j}) = 1, \tau^i \in \testset\}$ to be the set of instances $(i, j)$ where the precondition of action $a$ is satisfied at time-step $j$ of test trajectory $\tau^i \in \testset$.
We define precision and recall metrics for a particular action $a$ as in \Cref{eq:prec_rec}, where $\cpred, \cgt$ respectively correspond to the predicted and ground-truth preconditions.
The $F_1$ score is defined as the harmonic mean of precision and recall.
These metrics are macro-averaged across all actions $a\in\act$ to obtain the final metrics.
\begin{equation}
\text{Prec} = \frac{|\cpred \cap \cgt|}{|\cpred|}, \text{Rec} = \frac{|\cpred \cap \cgt|}{|\cgt|} 
\label{eq:prec_rec} %
\end{equation}
\vspace{-1.0em}
\paragraph{Baselines.}
We first consider a binary classifier which we call the \emph{Neural Precondition} baseline.
If we have labeled data of the form $(\ctxt, a, l)$, where $l \in \{0, 1\}$ indicates whether action $a$ is plausible given context $\ctxt$, we can train a binary classifier $h_a$ that predicts if the precondition for $a$ is satisfied.
However, in our setting, we only have positive instances where we know that the precondition of an action was satisfied, i.e., $\mathcal{D}^\text{pos} = \{(\ctxt, a_t, 1) | \tau \in \dataset, t \leq n\}$ where $\dataset$ is a dataset of expert trajectories.
In order to construct negative instances, we assume that the precondition for every future action $a_{t'} ; t' > t$ that appeared in the trajectory $\tau$ apart from the ground-truth action $a_t$ was not satisfied.
Based on this assumption we construct negative instances $\mathcal{D}^\text{neg} = \{(\ctxt, a_{t'}, 0) | t' > t, a_{t'} \neq a_t, \tau \in \dataset\}$ and train a binary classifier $h$ on $\mathcal{D}^\text{pos} \cup \mathcal{D}^\text{neg}$.
In practice, we uniformly sample instances from $\mathcal{D}^\text{pos}$ and $\mathcal{D}^\text{neg}$ to train the classifier.
We also consider a \emph{Prompting} baseline where we prompt a code model with a partial program specification and predict assertion statements with greedy decoding.

\begin{table}[!t]
\begin{adjustbox}{valign=t}
  \begin{minipage}[t]{0.56\textwidth}
    \centering
    \setlength{\tabcolsep}{3.8pt}
    \begin{tabular}{l c c c c c c}
    \toprule
    \multirow{2}{*}{Model}   & \multicolumn{3}{c}{SGD} & \multicolumn{3}{c}{Alfworld} \\
    \cmidrule(lr){2-4}
    \cmidrule(lr){5-7}
    & Prec & Rec & $F_1$ & Prec & Rec & $F_1$ \\
    \midrule
    Neural Precond   & 0.77 & 0.65 & 0.70 &  0.64 & 0.42 & 0.51 \\
    Prompting(2B)    & 0.73 & 0.75 & 0.74 &  0.59 & \textbf{0.89} & 0.71 \\
    Prompting(16B)   & 0.73 & 0.77 & 0.75 &  0.69 & 0.68 & 0.68 \\
    \midrule
    Ours(2B)       & 0.91 & 0.69 & 0.78 & \textbf{1.0} & 0.85 & \textbf{0.92} \\
    Ours(16B)      & \textbf{0.92} & \textbf{0.78} & \textbf{0.84} & \textbf{1.0} & 0.81 & 0.90 \\
    \bottomrule
    \end{tabular}
    \cuttableabove
    \caption{
    Precondition Inference performance on the SGD and Alfworld benchmarks.
    Metrics are precision, recall and $F_1$ score (All metrics higher $\uparrow$ the better).
    The simple prompting baseline and our approach are both evaluated with the \codegen 2B model and \starcoder 16B model.
    }
    \label{table:precond}
  \end{minipage}%
\end{adjustbox}
  \hfill
\begin{adjustbox}{valign=t}
  \begin{minipage}[t]{0.42\textwidth}
    \centering
    \setlength{\tabcolsep}{3.5pt}
    \begin{tabular}{l c c c c}
    \toprule
    \multirow{2}{*}{Approach} & \multicolumn{2}{c}{SGD} & \multicolumn{2}{c}{Alfworld} \\
    \cmidrule(lr){2-3} \cmidrule(lr){4-5}
    & $F_1$ & Cmp.  & SR & Cmp. \\
    \midrule
    BC          & 0.88 & 0.96 & 0.11 & 0.53 \\
    BC+Ours     & \textbf{0.89} & \textbf{0.98} & 0.14 & 0.74 \\
    \midrule
    Act         & 0.84 & 0.92 & 0.06 & 0.56 \\
    Act+Ours    & 0.85 & 0.97 & 0.23 & \textbf{0.96} \\
    \midrule
    ReAct       & - & - & 0.44 & 0.81 \\
    ReAct+Ours  & - & - & \textbf{0.48} & \textbf{0.96} \\
    \bottomrule
    \end{tabular}
    \caption{
    Policy performance on SGD and Alfworld benchmarks. 
    The performance metrics are $F_1$ score, task success rate (SR) and precondition compatibility (Cmp.) (All metrics higher $\uparrow$ the better).
    }
    \label{table:policy}

  \end{minipage}
\end{adjustbox}
\end{table}

\paragraph{Results.}

In \Cref{table:precond} we compare the precondition inference performance of different methods.
The precision for our models are high since the ranking criteria we adopted encourages high precision solutions. 
For example, in the \emph{Restaurants} domain of the SGD dataset, the ground truth precondition for the action \texttt{INFORM(slot)} is \texttt{requested\_slot[slot]}.\footnote{\texttt{`self'} and \texttt{`user'} omitted for brevity}
However, the predicted precondition is \texttt{(requested\_slot[slot] and query\_success==True)}.
The models pick up the fact that \texttt{query\_success==True} in all instances the \texttt{INFORM(slot)} action appeared in the demonstration trajectories. %
In this case precision is $1.0$ since the ground truth precondition will be satisfied whenever the predicted precondition is satisfied.
However, the recall is lower ($0.57$) since the predicted and ground-truth preconditions do not agree whenever \texttt{query\_success}$!=$\texttt{True}.

The Neural Precondition baseline generally underperforms other methods.
The assumption that actions different from the observed action for a given time-step in a trajectory are invalid does not always hold.
This is particularly an issue for the Alfworld benchmark where there can be multiple navigation actions in a trajectory while the agent is searching for an object which are equally plausible.
Comparing the labels in the generated data (e.g., $\mathcal{D}_\text{pos}, \mathcal{D}_\text{neg}$) with the corresponding ground-truth labels (computed using ground-truth preconditions), we observe that the (precision, recall) of the generated labels are respectively (1, 0.74) and (1, 0.60) for the SGD and Alfworld datasets.\footnote{Precision is 1.0 since labels in $\mathcal{D}_\text{pos}$ are guaranteed to be correct. Recall is lower since this is not the case for $\mathcal{D}_\text{neg}$.} 
When ground-truth precondition labels are used for training the classifier, performance improves (F1 scores of 0.81 and 0.67 in SGD and Alfworld, respectively), but generalization to unseen scenarios is still limited.

\Cref{table:qualitative} presents qualitative prediction results for all action types in the \emph{Buses} domain of the SGD benchmarks. 
Note that our models receive no supervision about preconditions and predict these candidates by only observing expert demonstrations.
We present qualitative prediction results for the Alfworld benchmark in \Cref{appsec:alfworld qual}.
Next, we analyze how these inferred action precondition can help build better agent policies.

\subsection{Precondition-aware Agent Policy}

\paragraph{Metrics.} 
In the SGD benchmark, we evaluate policy performance using actions in the demonstrations as reference.
Since the agent needs to predict multiple actions in a turn, we compute $F_1$ score treating demonstration actions as ground-truth.
For the embodied textworld task, a simulation environment is available, and we define the \textbf{success rate (SR)} metric which measures how often the agent successfully completed the given task.
We also define \textbf{precondition compatibility}, which measures how often the predicted action is compatible with (i.e., does not violate) the ground-truth preconditions.

\paragraph{Baselines.}
We consider the following baseline policy approaches.
\begin{itemize}[leftmargin=*]
\item \textbf{BC}: We consider a behavioral cloning baseline where a code model is fine-tuned (with LoRA) on the given expert demonstrations with a language modeling training objective.
\item \textbf{Act} \citep{yao2022react}: A few-shot prompting baseline where training demonstrations are provided as a prompt to a pre-trained model and the model predicts the next action. 
\item \textbf{ReAct} \citep{yao2022react}: A variation of the few-shot prompting strategy where the agent predicts a chain-of-thought natural language rationale before predicting each action. We design `think’ prompts similar to the original work and insert them as code comments in the demonstrations. At every step the agent generates an optional code comment (`think’ step) and predicts an action.
\end{itemize}

The \starcoder 16B model is used as the backbone model for all the policy methods unless specified otherwise.
We use preconditions predicted by the \starcoder 16B model in all experiments.

\paragraph{Results.}
We present the main results in \Cref{table:policy}.
The table shows the performance of each baseline policy (BC, Act, ReAct) as well as the performance when each of these methods are augmented with preconditions inferred by our approach.
We observe that the precondition-aware action sampling strategy combined with predicted preconditions lead to consistent performance improvements over each baseline policy.
In particular, we observe that the proposed precondition-based reasoning approach helps models generate actions that are more accurate and consistent with ground-truth preconditions.

On the Alfworld benchmark, our approach improves the success rate of the React agent from 44\% to 48\% and its precondition compatibility from 81\% to 96\%. %
This shows the synergistic potential of our approach and recent advances in prompting for policy learning problems such as generating natural language rationales before predicting actions.

These results shows that explicitly reasoning about preconditions helps build better policies.
Although few-shot prompting enables language/code models to perform tasks with limited supervision, they are generally limited by the number of demonstrations that can fit in the context window.
Condensing the information in multiple trajectories in a small set of rules (e.g., preconditions) can help overcome this limitation, and this idea may further be applicable to other tasks in general.

\subsection{Ablations and Analysis}

We perform a series of ablations to understand the impact of different components in our pipeline. 

\paragraph{How to incorporate precondition information?}
In \Cref{table:policy_ablation} we analyze the impact of (i) including precondition information in the prompt and (ii) the precondition-aware action sampling strategy.
Overall, while the inclusion of precondition information in the model prompt generally helps, our proposed precondition-aware action prediction strategy yields more consistent and significant improvements (e.g., $0.64$ to $0.68$ with predicted preconditions and $0.73$ with ground-truth preconditions in the 1-shot setting).
In addition, it helps predict actions that are more consistent with their preconditions (e.g., precondition compatibility improves to $>0.9$ for both 2B and 16B models).

\begin{figure}[!t]
\begin{adjustbox}{valign=t}
  \begin{minipage}[b]{0.53\textwidth}
    \setlength{\tabcolsep}{2.2pt}
    \centering
    \begin{tabular}{c c c c c c c}
    \toprule
    \multirowcell{2}{Model \\ size} & \multirowcell{2}{Prec. \\ prompt} & \multirowcell{2}{Prec. \\ sample} & \multicolumn{2}{c}{1-shot} & \multicolumn{2}{c}{10-shot} \\
    \cmidrule(lr){4-5} \cmidrule(lr){6-7}
    & & & $F_1\uparrow$ & Cmp.$\uparrow$ & $F_1\uparrow$ & Cmp.$\uparrow$ \\
    \midrule
    
    2B	& \xmark & \xmark & 0.51 & 0.73 & - & - \\
    2B	& \cmark & \xmark & 0.52 & 0.75 & - & - \\
    2B	& \cmark & \cmark & 0.58 & \textbf{0.92} & - & - \\
    \midrule
    16B	& \xmark & \xmark & 0.64 & 0.78 & 0.89 & 0.97 \\
    16B	& \cmark & \xmark & 0.65 & 0.77 & 0.90 & 0.97 \\
    16B	& \cmark & \cmark & \textbf{0.68} & 0.90 & \textbf{0.91} & \textbf{0.99} \\
    \bottomrule
    \end{tabular}
    \caption{
    Ablation to study the effect of (a) providing preconditions as part of the prompt (column 2) and (b) precondition-aware action prediction strategy (column 3) in the SGD benchmark.
    The 2B model can only accommodate upto 4 demonstrations and hence is not evaluated in the 10-shot setting.
    The performance metrics are $F_1$ score and precondition compatibility (Cmp.).
    }
    \label{table:policy_ablation}
  \end{minipage}
\end{adjustbox}
  \hfill
\begin{adjustbox}{valign=t}
  \begin{minipage}[b]{0.44\textwidth}
    \centering
    \begin{tikzpicture}
      \begin{axis}[ 
      width=\linewidth,
      line width=0.5,
      tick label style={font=\scriptsize},
      axis lines=left,
      xtick={2,4,6,8,10},
      ytick={0.5,0.6,0.7,0.8,0.9},
      xmin=0,
      xmax=11,
      ymin=0.5,
      ymax=0.95,
      label style={font=\small},
      legend style={draw=none, font=\scriptsize, align=right, xshift=0.4cm, yshift=0.4cm},
      legend pos=south east,
      xlabel={Num demos},
      x label style={at={(1.0,0.15)}},
      ylabel={F1},
      y label style={rotate=-90,at={(0.15,0.95)}},
      legend cell align={right}
      ]
    
    \addplot[blue,mark=*,mark size=0.5pt] coordinates {
    (1, 0.644508021666798)	(2, 0.752019159283598)	(3, 0.788461895290513)	(4, 0.840190528067724)	(5, 0.863297487847138)	(6, 0.884067388969947)	(7, 0.888793108245794)	(8, 0.889073431392768)	(9, 0.887703469839636)	(10, 0.892638492242011)
    };
    \addlegendentry{No precond. (16B)}
    
    \addplot[blue, dashed, mark=*, mark size=0.5pt] coordinates {
    (1, 0.676530800555165)	(2, 0.772204437369573)	(3, 0.807483713513961)	(4, 0.853358293672212)	(5, 0.869561139246354)	(6, 0.888385590298595)	(7, 0.893870163230869)	(8, 0.889836933665203)	(9, 0.896493704984293)	(10, 0.903395351044924)
    };
    \addlegendentry{Pred precond. (16B)}
    
    \addplot[blue, dotted, mark=*, mark size=0.5pt] coordinates {
    (1, 0.730282632744979)	(2, 0.796821530455961)	(3, 0.837449786218653)	(4, 0.867885001027201)	(5, 0.878857913573022)	(6, 0.898557159685443)	(7, 0.904451925658086)	(8, 0.899948644896883)	(9, 0.905597874321157)	(10, 0.905615917868757)
    };
    \addlegendentry{GT precond. (16B)}
    
    \addplot[red, mark=triangle*, mark size=1pt] coordinates {
    (1, 0.507177963771954)	(2, 0.642749365830165)	(3, 0.679228969388305)	(4, 0.747200986237821)						
    };
    \addlegendentry{No precond. (2B)~~}
    
    \addplot[red, dashed, mark=triangle*, mark size=1pt] coordinates {
    (1, 0.577216671250485)	(2, 0.671196170435615)	(3, 0.702588914494391)	(4, 0.771891204260681)
    };
    \addlegendentry{Pred precond. (2B)~~}
    
    \addplot[red, dotted, mark=triangle*, mark size=1pt] coordinates {
    (1, 0.577473247858972)	(2, 0.705870458888571)	(3, 0.728199631652704)	(4, 0.795079060335163)						
    };
    \addlegendentry{GT precond. (2B)~~}
    
    \end{axis}
    \end{tikzpicture}
    \caption{
    Ablation showing policy performance (F1 score) in the SGD benchmark when varying (a) the number of demonstrations (from 1 to 10), (b) model scale (\codegen 2B vs \starcoder 16B) and (c) precondition information available to the policy model (None/Predicted/Ground-truth).  
    }
    \label{fig:ablation}
    
  \end{minipage}%
\end{adjustbox}
\end{figure}

\paragraph{Amount of Supervision.}
\Cref{fig:ablation} shows model performance for varying amounts of supervision.
Due to it's maximum context size of 2048 tokens, the \codegen 2B model can only accommodate upto 4 demonstrations. %
Knowledge about preconditions is particularly helpful when the number of demonstrations is small.
Even as we increase the number of demonstrations, models continue to benefit from explicitly reasoning about preconditions.
Furthermore, performance with ground-truth preconditions shows that improvements in quality of preconditions lead to improvements in policy performance.

\paragraph{Model scale.}
We observe that the small model benefits more from precondition knowledge compared to the big model regardless of the amount of supervision.
Enhancing the reasoning capabilities of small models is important as big models demand higher costs and computation.
Leveraging precondition information and execution-based verification is a promising strategy to enhance small models. %

\section{Related Work}

\paragraph{Programs as Policies.}
There exist prior work that advocate viewing robot policies as code/programs \citep{liang2022code,singh2022progprompt}.
This is in contrast to most prior work that reason about plans almost entirely in natural language. 
Similar to natural language based prompting, LLMs are prompted with examples of programs corresponding to example tasks and are required to generate programs for a query task.
The programs can be rich and composed of functions supported by the target robot API or third party library functions.
Code comments help break down high-level task into subtasks and assertions are used to take environment feedback into account and provide an error recovery mechanism.
These prior work assume that precondition information is specified as part of the prompt.
In contrast, our work attempts to discover such information from action trajectories.
\paragraph{Reasoning with Verification.}
Prior work has attempted to augment capabilities of language models with programs and execution.
\citet{liu2022mind} combine LLMs with a physics simulator to answer physics questions.
Given a query, a text-to-code model trained with supervised learning generates a program in order to perform a simulation.
The simulator performs the simulation and produces an output, which is fed as additional information to the LLM to generate a response.
\citet{gao2022pal} interleave natural language chain-of-thought statements with program statements to perform calculations for arithmetic reasoning problems.
Analogous to these work, we find that the ability to verify via execution improves the performance of LMs, particularly small models. 
\paragraph{Reasoning via Prompting.}
Chain-of-thought prompting \citep{wei2022chain} has emerged as a powerful technique for getting language models to perform step-by-step reasoning.
These ideas have also been applied to planning problems \citep{huang2022inner,yao2022react} where agents are taught how to come up with rationales for predicted actions with few-shot demonstrations.
Compared to chain-of-thought rationales, which are dynamically generated on the fly and are problem-specific, we seek to identify a set of rules that capture action preconditions. 
This provides more controllability over the action generation process and the rules can also be vetted/edited to achieve desired behavior.
\paragraph{Structured Prediction with Programs.}
Programs have been used as a representation for structured prediction tasks in NLP \citep{wang2022code4struct,madaan2022language,zhang2023causal}. %
Code models have been used by these work for modeling procedural real-world activities, also called the script generation problem.
They find that code models have strong capability to reason about event sequences with minimal supervision compared to language models.
\paragraph{Subtask Graph Framework.}
Subtask Graphs are a modeling framework proposed by \citet{sohn2018hierarchical, sohn2020meta} to learn subtask preconditions from demonstrations.
Preconditions are modeled as boolean expressions involving a pre-defined set of boolean subtask variables which represent whether a subtask has been completed or not.
An Inductive Logic Programming (ILP) algorithm is used to identify the optimal boolean expression.
This framework was further extended to model real-world procedural activities in \citet{jang2023multimodal,logeswaran2023unsupervised}.
In contrast to the use of boolean expressions as the class of functions to model preconditions, the program representation we consider has the flexibility to represent a broader set of scenarios.
We draw inspiration from these works to formulate and evaluate the precondition inference component in our approach.

\section{Conclusion}
This work presented a novel approach to reason about action preconditions using programs for learning agent policies in a sequential decision making setting.
We proposed to use programs as a representation of the agent's observations and actions and showed that precondition inference and action prediction can be formulated as code-completion problems.
By leveraging the strong priors of pre-trained code models, we proposed a novel approach to infer action preconditions from demonstration trajectories without any additional supervision.
With the predicted preconditions, our precondition-aware action prediction strategy enables the agent to predict actions that are consistent with the preconditions and lead to better task completion compared to baselines.
Our study opens an exciting new direction to reason about action preconditions by leveraging code models.

\bibliography{ref,anthology,custom}
\bibliographystyle{iclr2024_conference}

\appendix
\onecolumn
\section{Program Specification}
\label{appsec:spec}

We present the program specification for the \schema{Restaurants} domain in the SGD benchmark in \Cref{fig:sgd prompt} and the Alfworld benchmark in \Cref{fig:alfworld prompt}.

\begin{figure*}[!h]
\begin{minipage}[t]{0.5\textwidth}
\scriptsize
\begin{minted}{python}
from collections import defaultdict

class USER:

  def __init__(self):
    self.informed_intent = defaultdict(lambda: False)
    self.informed_slot = defaultdict(lambda: False)
    self.requested_slot = defaultdict(lambda: False)

    self.no_more = False
    self.selected = False
    self.affirmed = False
    self.affirm_intent = False
    self.negate_intent = False
    self.request_alternatives = False

  def INFORM_INTENT(self, intent):
    self.informed_intent[intent] = True

  def NEGATE_INTENT(self):
    self.negate_intent = True

  def AFFIRM_INTENT(self):
    self.affirm_intent = True

  def REQUEST_ALTS(self):
    self.request_alternatives = True

  def INFORM(self, slot):
    self.informed_slot[slot] = True

  def REQUEST(self, slot):
    self.requested_slot[slot] = True

  def GOODBYE(self):
    self.no_more = True

  def THANK_YOU(self):
    self.no_more = True

  def SELECT(self):
    self.selected = True

  def AFFIRM(self):
    self.affirmed = True

  def NEGATE(self):
    self.affirmed = False


\end{minted}
\end{minipage}
\hfill
\begin{minipage}[t]{0.51\textwidth}
\scriptsize
\begin{minted}{python}


class SYSTEM:

  def __init__(self, user):
    self.user = user
    self.query_success = None

  def INFORM(self, slot):
    assert self.user.requested_slot[slot]

  def REQUEST(self, slot):
    assert not self.user.informed_slot[slot]

  def GOODBYE(self):
    assert self.user.no_more

  def FindRestaurants(self):
    assert self.user.informed_slot['city']
    assert self.user.informed_slot['cuisine']
    assert self.user.informed_intent['FindRestaurants']

  def ReserveRestaurant(self):
    assert self.user.selected or self.user.affirmed

  def OFFER(self, slot):
    assert self.query_success or self.user.affirmed

  def INFORM_COUNT(self):
    assert self.query_success

  def OFFER_INTENT(self, intent):
    assert self.user.selected

  def CONFIRM(self, slot):
    assert self.user.informed_slot['time']
    assert self.user.informed_slot['city']
    assert self.user.selected or \
      self.user.informed_slot['restaurant_name']

  def NOTIFY_SUCCESS(self):
    assert self.query_success

  def NOTIFY_FAILURE(self):
    assert not self.query_success

  def REQ_MORE(self):
    assert self.user.selected or self.user.no_more or \
      not self.query_success

  def set_query_status(self, status):
    self.query_success = status
\end{minted}
\end{minipage}
\caption{Program specification of the \schema{Restaurants} domain in the SGD benchmark. Note that the assertion statements are assumed to be unknown and only used for evaluation.}
\label{fig:sgd prompt}
\end{figure*}

\begin{figure*}[!h]
\centering
\begin{minipage}{\textwidth}
\centering
\scriptsize
\begin{minted}{python}
from collections import defaultdict

class Environment:

  def __init__(self):
    self.object_states = defaultdict(lambda: False)

  def set_property(self, obj, property, value):
    self.object_states[(obj, property)] = value

  def get_property(self, obj, property):
    return self.object_states[(obj, property)]

class Agent:

  def __init__(self, env):
    self.env = env
    self.inventory = None
    self.visible_objects = set()

  def add_inventory(self, obj):
    self.inventory = obj

  def remove_inventory(self, obj):
    self.inventory = None

  def update_visible_objects(self, *args):
    self.visible_objects.update(list(args))

  def is_visible(self, obj):
    return obj in self.visible_objects

  def goto(self, recep):
    assert self.is_visible(recep)

  def open(self, recep):
    assert self.is_visible(recep)
    assert self.env.get_property(recep, 'open') == False

  def close(self, recep):
    assert self.is_visible(recep)
    assert self.env.get_property(recep, 'open') == True

  def take(self, obj, recep):
    assert self.is_visible(obj)
    assert self.is_visible(recep)
    assert self.inventory == None, 'Can only hold one object at a time'

  def put(self, obj, recep):
    assert self.inventory == obj, 'Need to be holding the object'
    assert self.is_visible(recep)

  def clean(self, obj, recep):
    assert self.is_visible(recep)
    assert self.inventory == obj, 'Need to be holding the object'
    assert 'sink' in recep

  def heat(self, obj, recep):
    assert self.is_visible(recep)
    assert self.inventory == obj, 'Need to be holding the object'
    assert 'microwave' in recep

  def cool(self, obj, recep):
    assert self.is_visible(recep)
    assert self.inventory == obj, 'Need to be holding the object'
    assert 'fridge' in recep

  def toggle(self, obj):
    assert self.is_visible(obj)
\end{minted}
\end{minipage}
\caption{Program specification of the Alfworld benchmark. Note that the assertion statements are assumed to be unknown and only used for evaluation.}
\label{fig:alfworld prompt}
\end{figure*}

\clearpage
\section{Example Trajectories}
\label{appesec:examples}

\Cref{fig:sgd example} shows an example trajectory from the \schema{Restaurants} domain in the SGD benchmark and the corresponding program representation.
\Cref{fig:alfworld text,fig:alfworld code} show an example trajectory from the \schema{pick and place} task of the Alfworld benchmark in its original text representation and the corresponding program representation.

\begin{figure*}[!h]
\begin{minipage}{0.45\textwidth}
\scriptsize
\begin{lstlisting}[backgroundcolor = \color{white}]
USER:   (INFORM_INTENT, FindRestaurants)
SYSTEM: (REQUEST, city)
USER:   (INFORM, city)
SYSTEM: (REQUEST, cuisine)
USER:   (INFORM, cuisine)
SYSTEM: (FindRestaurants)
        (Query successful)
        (OFFER, restaurant_name)
        (OFFER, city)
        (INFORM_COUNT)
USER:   (REQUEST_ALTS)
SYSTEM: (OFFER, restaurant_name)
        (OFFER, city)
USER:   (REQUEST, has_live_music)
SYSTEM: (INFORM, has_live_music)
USER:   (INFORM_INTENT, ReserveRestaurant)
        (SELECT)
SYSTEM: (REQUEST, time)
USER:   (INFORM, time)
SYSTEM: (CONFIRM, restaurant_name)
        (CONFIRM, city)
        (CONFIRM, time)
        (CONFIRM, party_size)
        (CONFIRM, date)
USER:   (INFORM, date)
        (NEGATE)
SYSTEM: (CONFIRM, time)
        (CONFIRM, date)
USER:   (INFORM, time)
        (NEGATE)
SYSTEM: (CONFIRM, time)
        (CONFIRM, date)
USER:   (AFFIRM)
SYSTEM: (ReserveRestaurant)
        (Query successful)
        (NOTIFY_SUCCESS)
USER:   (REQUEST, serves_alcohol)
SYSTEM: (INFORM, serves_alcohol)
USER:   (THANK_YOU)
        (GOODBYE)
SYSTEM: (GOODBYE)
\end{lstlisting}
\end{minipage}
\hfill
\begin{minipage}{0.5\textwidth}
\scriptsize
\begin{minted}{python}
user.INFORM_INTENT('FindRestaurants')
system.REQUEST('city')
user.INFORM('city')
system.REQUEST('cuisine')
user.INFORM('cuisine')
system.FindRestaurants()
system.set_query_status(True)
system.OFFER('restaurant_name')
system.OFFER('city')
system.INFORM_COUNT()
user.REQUEST_ALTS()
system.OFFER('restaurant_name')
system.OFFER('city')
user.REQUEST('has_live_music')
system.INFORM('has_live_music')
user.INFORM_INTENT('ReserveRestaurant')
user.SELECT()
system.REQUEST('time')
user.INFORM('time')
system.CONFIRM('restaurant_name')
system.CONFIRM('city')
system.CONFIRM('time')
system.CONFIRM('party_size')
system.CONFIRM('date')
user.INFORM('date')
user.NEGATE()
system.CONFIRM('time')
system.CONFIRM('date')
user.INFORM('time')
user.NEGATE()
system.CONFIRM('time')
system.CONFIRM('date')
user.AFFIRM()
system.ReserveRestaurant()
system.set_query_status(True)
system.NOTIFY_SUCCESS()
user.REQUEST('serves_alcohol')
system.INFORM('serves_alcohol')
user.THANK_YOU()
user.GOODBYE()
system.GOODBYE()
\end{minted}
\end{minipage}
\caption{Example trajectory from the \schema{Restaurants} domain of the SGD benchmark. The original data representation is shown on the left and the corresponding program representation is shown on the right.}
\label{fig:sgd example}
\end{figure*}

\begin{figure*}[!h]
\centering
\begin{minipage}{\textwidth}
\centering
\scriptsize
\begin{lstlisting}[backgroundcolor = \color{white}]
You are in the middle of a room. Looking quickly around you, you see a countertop 1, a drawer 1, a drawer 2, a drawer 3, a drawer 4, a drawer 5, a drawer 6, a garbagecan 1, a handtowelholder 1, a handtowelholder 2, a sinkbasin 1, a toilet 1, a toiletpaperhanger 1, a towelholder 1, and a towelholder 2.
Your task is to: put some spraybottle on garbagecan
> go to drawer 4
The drawer 4 is closed.
> open drawer 4
You open the drawer 4. The drawer 4 is open. In it, you see nothing.
> close drawer 4
You close the drawer 4.
> go to drawer 2
The drawer 2 is closed.
> open drawer 2
You open the drawer 2. The drawer 2 is open. In it, you see nothing.
> close drawer 2
You close the drawer 2.
> go to drawer 3
The drawer 3 is closed.
> open drawer 3
You open the drawer 3. The drawer 3 is open. In it, you see nothing.
> close drawer 3
You close the drawer 3.
> go to countertop 1
On the countertop 1, you see a sink 1, a soapbottle 1, and a toiletpaper 1.
> go to toilet 1
On the toilet 1, you see a cloth 1, and a spraybottle 1.
> take spraybottle 1 from toilet 1
You pick up the spraybottle 1 from the toilet 1.
> go to garbagecan 1
On the garbagecan 1, you see nothing.
> put spraybottle 1 in/on garbagecan 1
You put the spraybottle 1 in/on the garbagecan 1.
\end{lstlisting}
\end{minipage}
\caption{Example trajectory from the \schema{pick and place} task of the Alfworld benchmark in its original text representation.}
\label{fig:alfworld text}
\end{figure*}

\begin{figure*}[!h]
\centering
\begin{minipage}{\textwidth}
\centering
\scriptsize
\begin{minted}{python}
def put_some_spraybottle_on_garbagecan():
    # put some spraybottle on garbagecan.
    env = Environment()
    agent = Agent(env)
    agent.update_visible_objects('countertop 1', 'drawer 1', 'drawer 2', 'drawer 3', 'drawer 4', 'drawer 5', \
      'drawer 6', 'garbagecan 1', 'handtowelholder 1', 'handtowelholder 2', 'sinkbasin 1', 'toilet 1', \
      'toiletpaperhanger 1', 'towelholder 1', 'towelholder 2')
    agent.goto('drawer 4')
    env.set_property('drawer 4', 'open', False)
    agent.open('drawer 4')
    env.set_property('drawer 4', 'open', True)
    agent.close('drawer 4')
    env.set_property('drawer 4', 'open', False)
    agent.goto('drawer 2')
    env.set_property('drawer 2', 'open', False)
    agent.open('drawer 2')
    env.set_property('drawer 2', 'open', True)
    agent.close('drawer 2')
    env.set_property('drawer 2', 'open', False)
    agent.goto('drawer 3')
    env.set_property('drawer 3', 'open', False)
    agent.open('drawer 3')
    env.set_property('drawer 3', 'open', True)
    agent.close('drawer 3')
    env.set_property('drawer 3', 'open', False)
    agent.goto('countertop 1')
    agent.update_visible_objects('countertop 1', 'sink 1', 'soapbottle 1', 'toiletpaper 1')
    agent.goto('toilet 1')
    agent.update_visible_objects('toilet 1', 'cloth 1', 'spraybottle 1')
    agent.take('spraybottle 1', 'toilet 1')
    agent.add_inventory('spraybottle 1')
    agent.goto('garbagecan 1')
    agent.update_visible_objects('garbagecan 1')
    agent.put('spraybottle 1', 'garbagecan 1')
    agent.remove_inventory('spraybottle 1')

\end{minted}
\end{minipage}
\caption{Example trajectory from the \schema{pick and place} task of the Alfworld benchmark in our program representation.}
\label{fig:alfworld code}
\end{figure*}

\clearpage
\section{Qualitative Prediction Results for Alfworld}
\label{appsec:alfworld qual}
\Cref{table:alfworld qualitative} shows precondition prediction results for the Alfworld benchmark.

\begin{figure}[!h]
\begin{minipage}[t]{0.5\textwidth}
\begin{minted}
[
framesep=2mm,
fontsize=\scriptsize,
]
{python}
# Ground-Truth Preconditions                          

def goto(self, recep):
  assert self.is_visible(recep)

def open(self, recep):
  assert self.is_visible(recep)
  assert self.env.get_property(recep, 'open') == False

def close(self, recep):
  assert self.is_visible(recep)
  assert self.env.get_property(recep, 'open') == True

def take(self, obj, recep):
  assert self.is_visible(obj)
  assert self.is_visible(recep)
  assert self.inventory == None

def put(self, obj, recep):
  assert self.inventory == obj
  assert self.is_visible(recep)

def clean(self, obj, recep):
  assert self.is_visible(recep)
  assert self.inventory == obj
  assert 'sink' in recep

def heat(self, obj, recep):
  assert self.is_visible(recep)
  assert self.inventory == obj
  assert 'microwave' in recep

def cool(self, obj, recep):
  assert self.is_visible(recep)
  assert self.inventory == obj
  assert 'fridge' in recep

def toggle(self, obj):
  assert self.is_visible(obj)
\end{minted}
\end{minipage}
\hfill
\begin{minipage}[t]{0.5\textwidth}
\begin{minted}
[
framesep=2mm,
fontsize=\scriptsize,
]
{python}
# Predicted Preconditions                          

 def goto(self, recep):
  assert self.is_visible(recep)

def open(self, recep):
  assert recep in self.visible_objects
  assert self.env.get_property(recep, 'open') is False

def close(self, recep):
  assert self.env.get_property(recep, 'open')


def take(self, obj, recep):
  assert self.is_visible(obj)
  assert self.inventory is None


def put(self, obj, recep):
  assert obj in self.inventory


def clean(self, obj, recep):
  assert obj in self.inventory



def heat(self, obj, recep):
  assert self.inventory != None
  assert self.env.get_property(obj, 'heat') == False


def cool(self, obj, recep):
  assert obj in self.inventory
  assert self.env.get_property(obj, 'cool') == False


def toggle(self, obj):
  assert self.is_visible(obj)    
  assert self.inventory is not None
\end{minted}
\end{minipage}
\caption{Ground-truth and predicted preconditions for actions in the Alfworld benchmark.}
\label{table:alfworld qualitative}
\end{figure}

\clearpage

\section{Precondition Inference}

\begin{figure*}[!h]
\centering
\includegraphics[width=\textwidth]{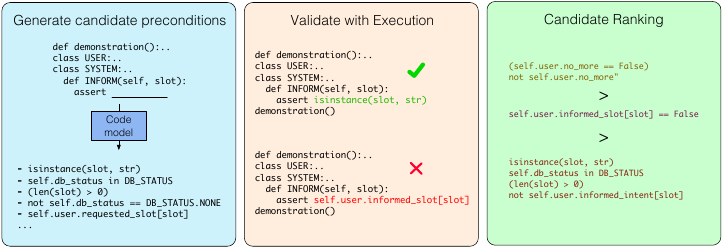}
\cutfigureabove
\caption{\textbf{Precondition Inference Overview}: 
We first generate multiple precondition candidates by prompting a code model with demonstration programs. 
We then validate these candidates based on the demonstrations via execution, and rank them to identify the most promising candidates.
}
\label{fig:precond}
\cutfigurebelow
\end{figure*}

\Cref{fig:precond} illustrates the pipeline for action precondition generation.
\label{appsec:precondition inference}

\section{Precondition-aware Action Prediction}
\label{appsec:action prediction}

\Cref{alg:sampling} presents the pseudocode for our precondition-aware action sampling strategy.

\begin{algorithm}[!h]
\small
\caption{Sample Next Action with Verification}
\label{alg:sampling}
\begin{algorithmic}[0]
    \State \textbf{Inputs}: \texttt{query}, \texttt{max\_attempts}
    \State $\texttt{verified} \gets \textit{False}$
    \State $\texttt{attempts} \gets 0$
    \State $\texttt{prediction} \gets \Call{GreedySample}{$\texttt{query}$}$
    \While{not \texttt{verified} and \texttt{attempts} < \texttt{max\_attempts}}
        \If{$\texttt{attempts} > 0$}
            \State $\texttt{prediction} \gets \Call{RandomSample}{$\texttt{query}$}$
        \EndIf
        \State $\texttt{program} \gets \texttt{query} + \texttt{prediction}$
        \State $\texttt{verified} \gets \Call{VerifyProgram}{$\texttt{program}$}$
        \State $\texttt{attempts} \gets \texttt{attempts} + 1$
    \EndWhile
    \If{not \texttt{verified}}
        \State $\texttt{prediction} \gets \Call{GreedySample}{$\texttt{query}$}$
    \EndIf
    \State \textbf{Outputs}: \texttt{prediction}, \texttt{verified}
\end{algorithmic}
\end{algorithm}

\clearpage

\section{Precondition Inference Example}
\label{appsec:precond inf ex}

Below we present an example showing the intermediate outputs of the precondition inference pipeline for action \texttt{INFORM} in the \schema{Restaurants} domain of the SGD benchmark.
The \texttt{INFORM} action takes a slot argument and the function syntax is \texttt{INFORM(self, slot)} (See the full program specification in \Cref{appsec:spec}).

\begin{itemize}[leftmargin=*]
\item Generate candidate preconditions $\mathcal{H}_a^\text{initial}$ by prompting a code model with demonstrations
\begin{minted}{python}
    not self.user.requested_slot[slot], 'Requested slot'
    self.query_success!= None
    (self.query_success == True)
    (self.query_success)
    self.query_success is not None
    (self.user.informed_slot[slot])
    self.query_success is None
    (slot in self.user.informed_slot)
    slot not in self.user.requested_slot.keys()
    self.user.affirmed == True
    self.user.affirmed
    (hasattr(slot, '__name__'))
    isinstance(slot, str)
    slot!= 'date'
    self.query_success in (True, False)
    slot!='serves_alcohol'
    self.user.requested_slot[slot]
\end{minted}

\item Identify valid candidates $\mathcal{H}_a^\text{valid}$ based on execution against the demonstration programs
\begin{minted}{python}
    self.query_success!= None
    (self.query_success == True)
    (self.query_success)
    self.query_success is not None
    isinstance(slot, str)
    slot!= 'date'
    self.query_success in (True, False)
    self.user.requested_slot[slot]
\end{minted}

\item Identify candidates that are functionally equivalent
\begin{minted}{python}
    # Cluster 0
    self.query_success!= None
    self.query_success is not None
    self.query_success in (True, False)
    
    # Cluster 1
    (self.query_success == True)
    (self.query_success)
    
    # Cluster 2
    isinstance(slot, str)
    slot!= 'date'
    
    # Cluster 3
    self.user.requested_slot[slot]
\end{minted}

\item Identify the precondition clusters that satisfy \Cref{eq:hopt}
\begin{minted}{python}
    # Cluster 1 (subsumes both cluster 0 and 2)
    (self.query_success == True)
    (self.query_success)
    
    # Cluster 3
    self.user.requested_slot[slot]
\end{minted}

\item Choose single representative (randomly) from each cluster to construct $\mathcal{H}_a^\text{opt}$
\begin{minted}{python}
    (self.query_success == True)
    self.user.requested_slot[slot]
\end{minted}

\end{itemize}

\end{document}